\documentclass[11pt]{article}
\usepackage{amsmath}
\usepackage[final]{acl}

\usepackage{times}
\usepackage{latexsym}

\usepackage[T1]{fontenc}

\usepackage[utf8]{inputenc}

\usepackage{microtype}

\usepackage{inconsolata}

\usepackage{graphicx}
\usepackage{subcaption}

\usepackage{algorithm}
\usepackage{algpseudocode}
\usepackage{amsmath}
\usepackage{amssymb}
\usepackage{algorithm}
\usepackage{algpseudocode}
\usepackage{amsmath}
\usepackage{amssymb}
\usepackage{hyperref}

\algrenewcommand{\algorithmiccomment}[1]{\hspace{1em}\{#1\}}

\title{Beyond Memorization: Distinguishing Between Pattern-Based and Epistemic Reasoning in LLMs Using Epistemic Puzzles}

\author{{Adi Gabay, Gabriel Stanovsky, Liat Peterfreund}\\ 
{School of Computer Science and Engineering} \\
{The Hebrew University of Jerusalem}\\
adi.gabay1@mail.huji.ac.il}

\usepackage{booktabs}

\begin{document}
\maketitle
\begin{abstract}

Epistemic reasoning requires agents to infer the state of the world from partial observations and information about other agents' knowledge. Prior work evaluating LLMs on epistemic puzzles often frames failures as memorization rather than reasoning. We argue that this dichotomy is too coarse for newer models: memorization is a limiting case of \emph{pattern-based reasoning}, where a model matches a task to a familiar template and applies the corresponding solution. We introduce a two-dimensional benchmark over Dynamic Epistemic Logic (DEL) puzzles, separating narrative familiarity from inference complexity, which allows us to distinguish pattern-based from epistemic reasoning. 
We find that models are substantially more robust to surface form changes than prior work suggested, yet consistently struggle in asymmetric settings where familiar patterns no longer apply and success requires tracking fragmented epistemic states. Code, datasets, and evaluation results are available at \url{https://github.com/adigabay84/del-puzzles-benchmark}.

\end{abstract}

\section{Introduction}

Problems requiring epistemic reasoning in multi-agent systems involve inferring the state of the world by integrating one’s own observations with information about other agents’ knowledge~\cite{fagin1995}.
Such challenges arise naturally in collaborative settings where each agent has only a partial view of the world, such as legal proceedings, education, or medical applications~\cite{pmlr-v80-rabinowitz18a,li-etal-2023-theory}.

\begin{figure}[t!]
    \includegraphics[width=\columnwidth]
    {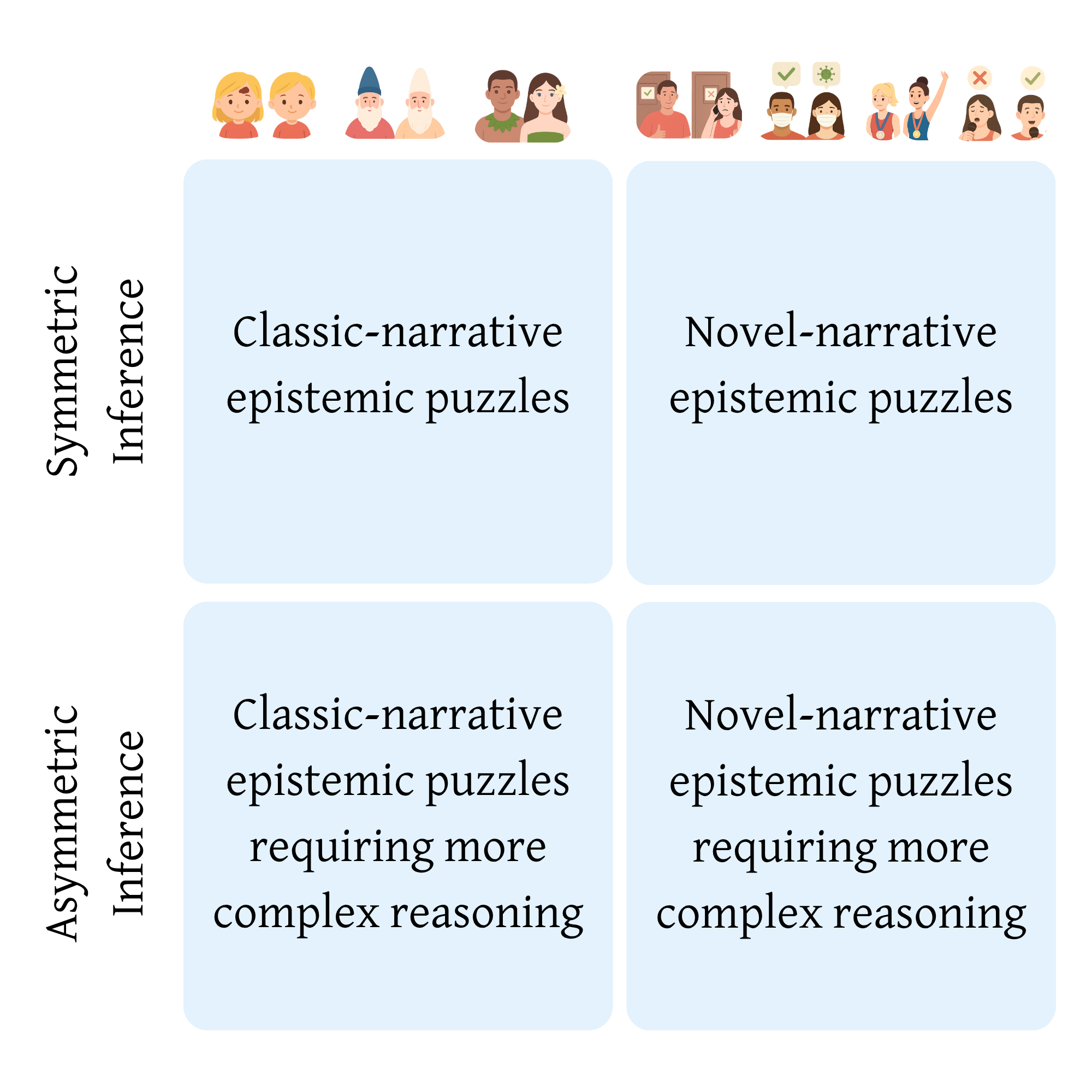}
    \caption{\textbf{The two-dimensional benchmark for distinguishing between pattern-based and epistemic reasoning in LLMs.}
    The horizontal Narrative dimension contrasts three classic puzzles (\textit{Muddy Children}, 
    \textit{Wise Men}, and \textit{Blue-Eyed Islanders}) with four novel ones (\textit{Safety Inspection}, \textit{Health Screening}, \textit{Olympic Games}, and \textit{Singing Contest}). The vertical Inference dimension increases reasoning difficulty by shifting from standard symmetric settings, where each agent observes all others but themselves, to asymmetric, partial views.}
    \label{fig:benchmark}
\end{figure}

Previous work evaluating LLMs' epistemic reasoning has focused on canonical problems such as the \textit{Muddy Children} problem, reporting limited performance~\cite{sileo2023}. Related benchmarks show performance degradation under task variations: when names and numbers are altered in arithmetic word problems~\cite{mirzadeh2025gsm}, when puzzle complexity increases~\cite{lin2025zebralogic,shojaee2026illusion}, and when non-standard variants change the underlying rules or constraints~\cite{liang2025hardcorelogic}. These findings are often framed as a contrast between \emph{reasoning}, where a model derives a solution from the specific parameters of the instance, and \emph{memorization}, where it matches the task to a familiar pretraining pattern and reproduces the associated answer. Under this view, models over-rely on memorization, which breaks down under superficial perturbations~\cite{jiang2024,mirzadeh2025gsm}.



In this work, we argue that recent advances in LLMs warrant a more nuanced treatment of memorization. Rather than viewing memorization as a binary alternative to reasoning, we treat it as one form of a broader strategy that we define as \emph{pattern-based reasoning}. Under this strategy, a model recognizes a problem as an instance or variant of a familiar template and applies the corresponding solution pattern. The central challenge is therefore not to reason directly over the specific  parameters of the instance, but to identify the familiar structure to which it can be mapped. Under this view, memorization is the limiting case in which the model has encountered the same instance previously.

We therefore ask: \emph{Can pattern-based and epistemic reasoning be distinguished in LLMs, and can performance under each be quantified?}
We study this question through a controlled family of Dynamic Epistemic Logic (DEL) puzzles, instantiating core epistemic mechanisms to isolate narrative familiarity from inference complexity.

Instead of seeking specific perturbations that “break” a model, we propose a two-dimensional matrix taxonomy (Figure~\ref{fig:benchmark}) that systematically disentangles pattern-based reasoning from epistemic reasoning. The horizontal dimension varies \emph{narrative familiarity}, spanning seven distinct narratives: three classic puzzles - \textit{Muddy Children} ~\cite{van2008dynamic}, \textit{Wise Men}~\cite{d2009neural} and \textit{Blue-Eyed Islanders}~\cite{stuhlmuller2014reasoning}, compared with four novel narratives we design specifically for this benchmark, and differ substantially from the familiar theme of classic puzzles - \textit{Safety Inspection}, \textit{Health Screening}, \textit{Olympic Games}, and \textit{Singing Contest}. These are designed to keep the exact reasoning needed to solve epistemic puzzles while changing the setting, entities, and surface narrative.

The vertical dimension scales \emph{inference complexity}, shifting from standard symmetric observations where each agent observes all others but themselves, to asymmetric observations where each agent observes a randomly sampled set of the other agents. Measuring performance across this matrix allows us to characterize how models behave as pattern-based reasoning becomes less reliable and success depends heavily on epistemic reasoning.

We evaluate various LLMs and find that while most models perform well in classic, symmetric settings, even when surface patterns are substantially modified, they struggle in asymmetric cases that require multi-step epistemic reasoning.
This drop reveals distinct behavioral profiles based on how models 
handle narrative familiarity and structural complexity. While some models show a relative advantage on classic narratives and others on novel ones, the remaining models show no meaningful difference between the two. Beyond the benchmark itself, this distinction provides an interpretable lens on model behavior, helping reveal not only how well models perform, 
but also \emph{how} their reasoning changes under different settings.

Our contributions are threefold: (1) we reinterpret “memorization” as a form of \emph{pattern-based reasoning}
 (2) we introduce a matrix evaluation framework that isolates narrative familiarity from inference complexity; and (3) we show that while models often succeed under conditions well-suited to pattern-based reasoning, they struggle with complex epistemic scenarios, exposing a spectrum of distinct behavioral profiles across current LLMs.


\begin{figure*}[t]
    \centering
    \begin{subfigure}[b]{0.32\textwidth}
        \centering
        \includegraphics[width=\textwidth]
        {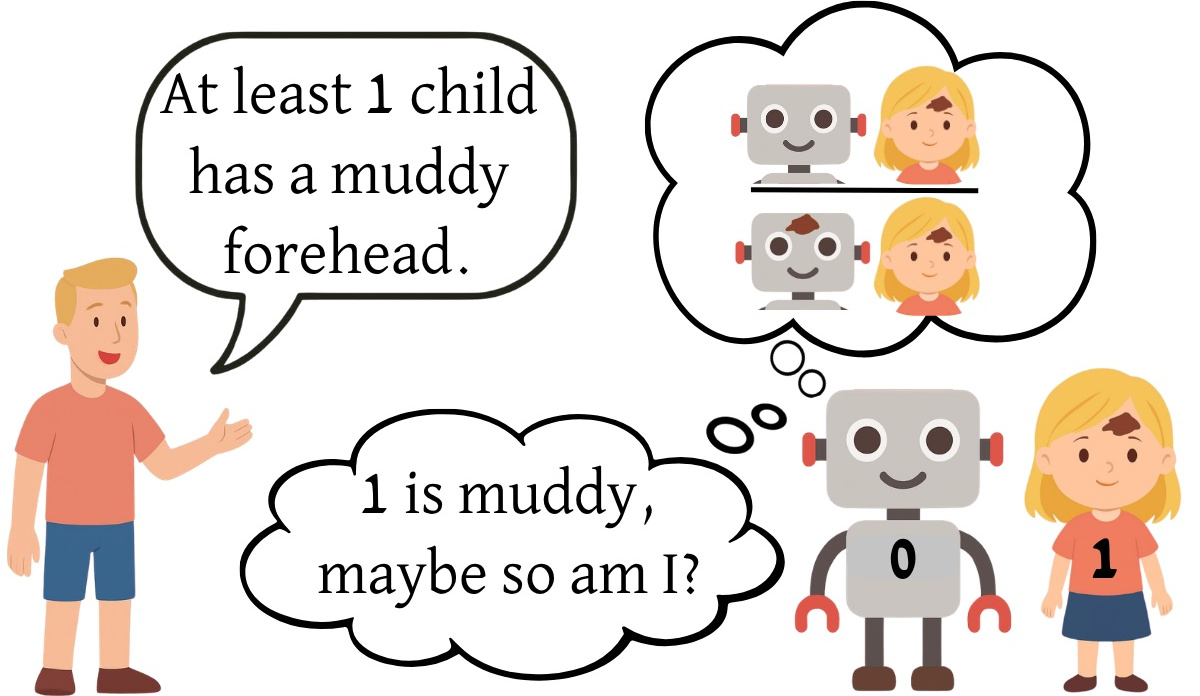}
        \caption{Public Announcement: Following the announcement, child 1 eliminates the possibility that both children are clean.}
        \label{fig:announcement}
    \end{subfigure}
    \hfill
    \begin{subfigure}[b]{0.32\textwidth}
        \centering
        \includegraphics[width=\textwidth]{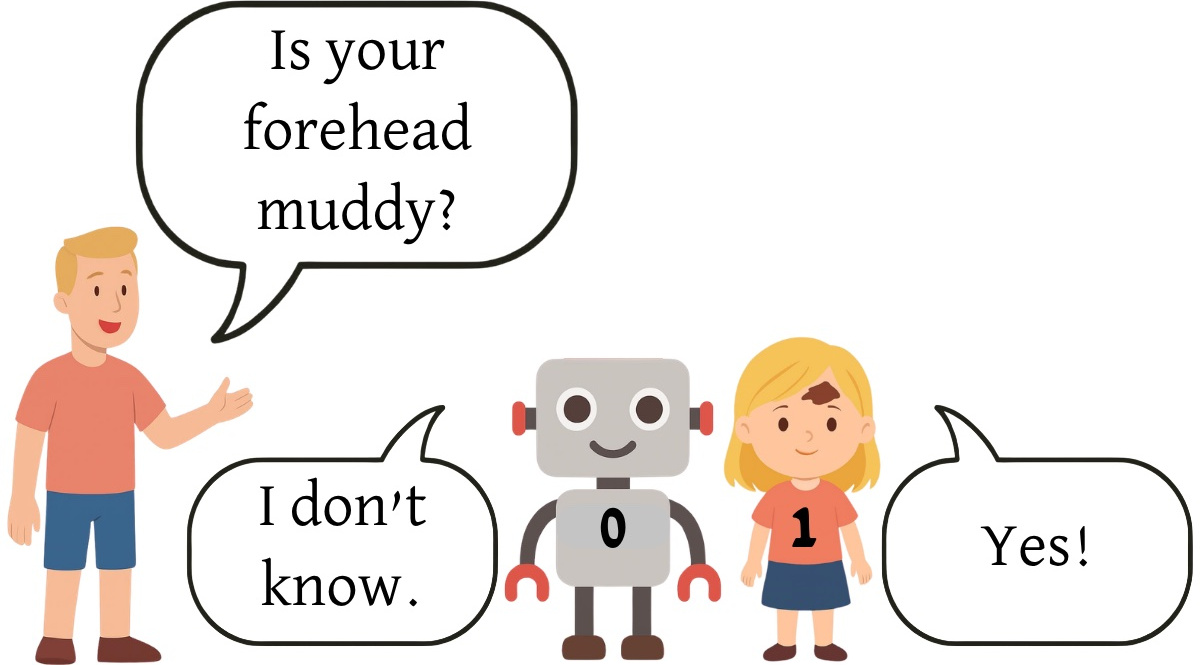}
        \caption{Round 1: Child 0 cannot rule out possibilities based on the public announcement, while child 1 knows she is muddy.}
        \label{fig:round1}
    \end{subfigure}
    \hfill
    \begin{subfigure}[b]{0.32\textwidth}
        \centering
        \includegraphics[width=\textwidth]{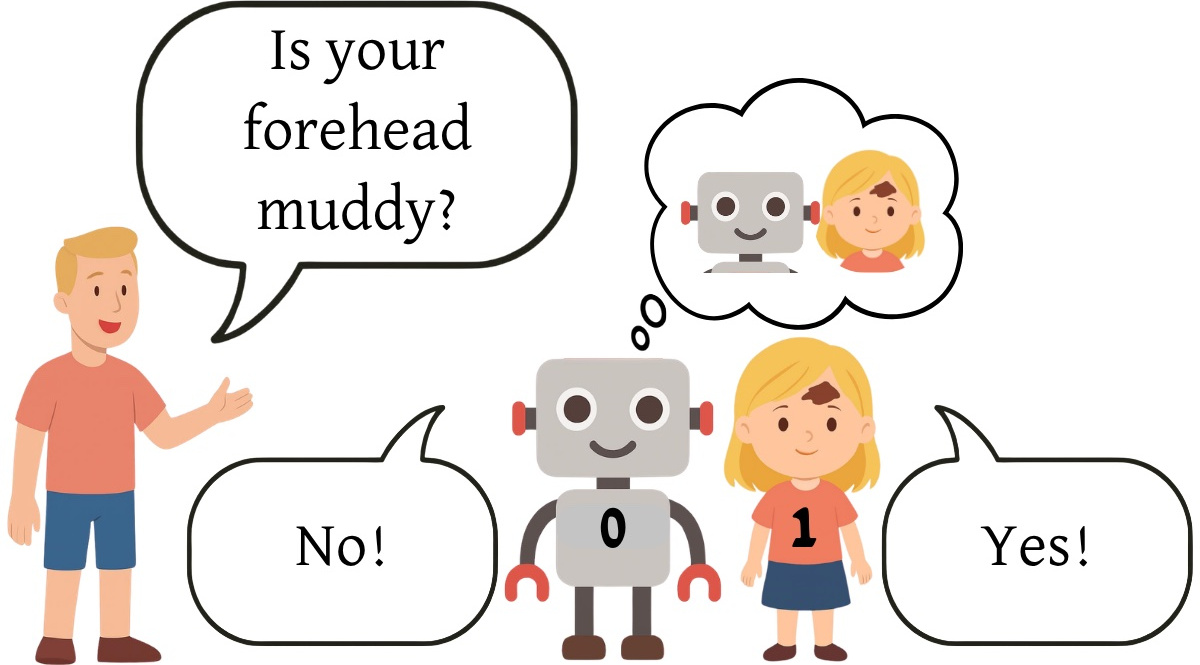}
        \caption{Round 2: Child 0 knows he is clean, since if he were muddy, child 1 would have remained uncertain.}
        \label{fig:round2}
    \end{subfigure}
    
    \caption{ 
    An example of the \textit{Muddy Children} puzzle with two children, one of whom is muddy, where each child observes the other child's forehead but not their own. Thought bubbles depict the possibilities child 0 considers.}

    \label{fig:reasoning_rounds}
\end{figure*}

\section{Background: Dynamic Epistemic Logic}

Dynamic Epistemic Logic (DEL) is a formal framework for modeling what agents know, which possibilities they consider, and how knowledge evolves as new information is revealed. In this framework, \emph{agents} are entities whose knowledge is being modeled, and \emph{public announcements} rule out possibilities inconsistent with newly available information, across multiple \emph{rounds} of communication~\cite{baltagMossSolecki1998,van2008dynamic}.

DEL is essential for multi-agent systems as widely emphasized in the literature ~\cite{bolander2011epistemic, belardinelli2009quantified, engesser2017cooperative}. It captures a distinct form of reasoning: not only deriving facts, but also tracking higher-order knowledge (i.e., knowledge about others' knowledge) and how these evolve as agents reason about each other under partial information.

We focus on canonical DEL puzzles as our baseline.
These puzzles are not merely examples, but compact tests of DEL's core mechanisms - knowledge inference, higher-order knowledge, common knowledge, and knowledge dynamics - requiring models to reason from partial observations, public updates, and others' uncertainty.

Figure~\ref{fig:reasoning_rounds} illustrates a two-child instance of the \textit{Muddy Children} puzzle. Each child can see the other's forehead but not their own, so their uncertainty is represented by the possible worlds in the thought bubbles. In panel~\ref{fig:announcement}, both remain uncertain about themselves; the father’s announcement that at least one child is muddy rules out the all-clean world. This lets child $1$ infer that she is muddy, and after she announces this in panel~\ref{fig:round1}, child $0$ can further narrow the possibilities and reach his own conclusion in panel~\ref{fig:round2}.



\section{Distinguishing Between Pattern-Based Reasoning and Epistemic Reasoning}

We aim to distinguish between two ways in which LLMs may determine the correct response of an agent in DEL-style settings: \emph{pattern-based reasoning} and \emph{epistemic reasoning}. Under pattern-based reasoning, the model identifies the epistemic state as a variant of a familiar problem and applies the corresponding solution. Under epistemic reasoning, by contrast, the model derives the agent’s response  from the observations, public announcements, and prior answers in the interaction rounds.


\subsection{Task Definition}

\begin{figure}[t]
    \centering
    \includegraphics[width=\columnwidth]{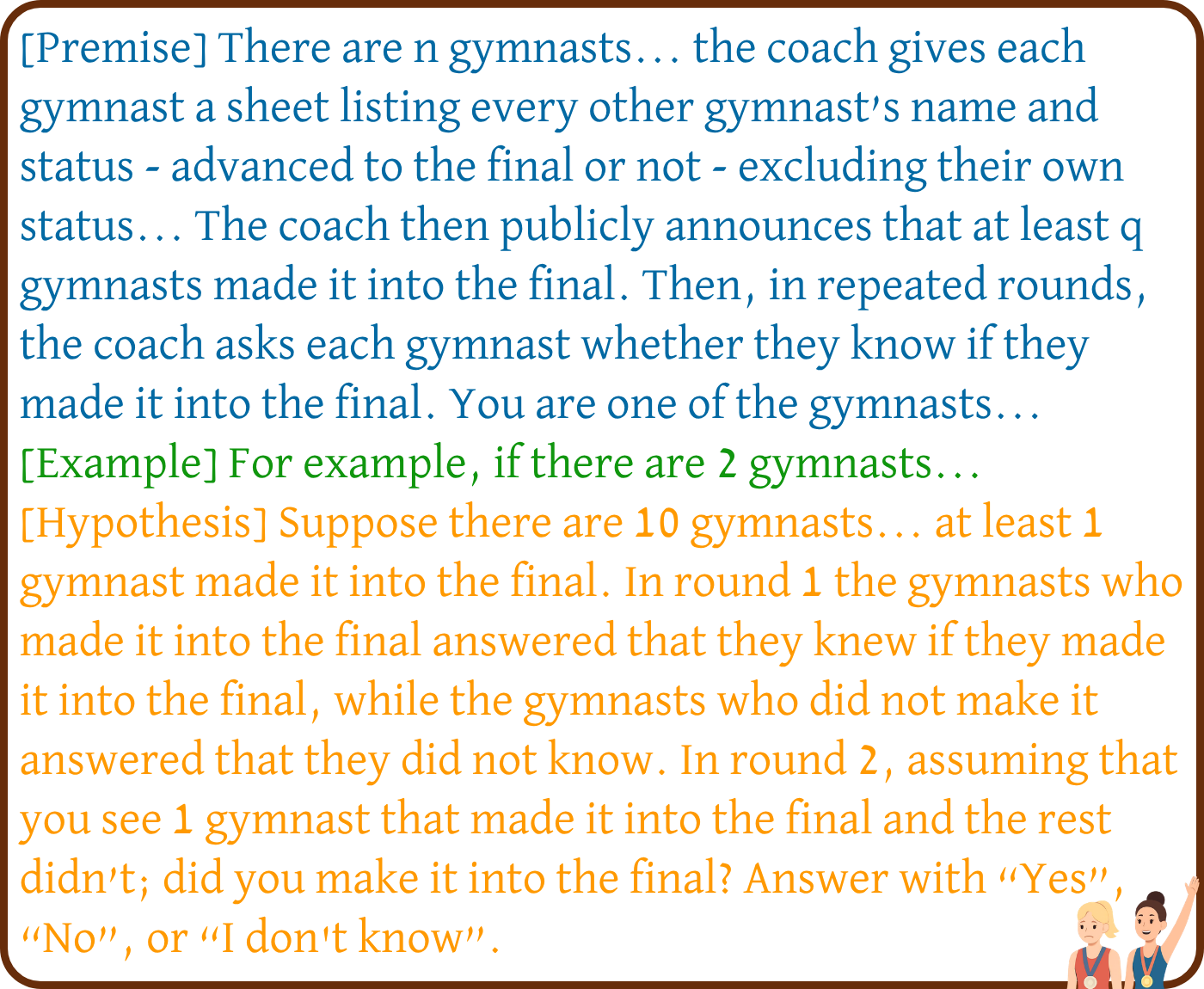}
    \caption{\textbf{Example of a reduced symmetric inference prompt in the \textit{Olympic Games} setup.}
    Each prompt contains the setup of the puzzle and the model's task (blue), a short example illustrating the required response format (green), and a test instance specifying the public interaction history and the agents' observations (orange).}
    \label{fig:prompt_template}
\end{figure}

To evaluate LLMs' epistemic reasoning ability, we place the LLM as an agent in a simulated world (e.g., a child in the \textit{Muddy Children} puzzle). The model's task is to infer its own status by reasoning over the information in its prompt, as illustrated by the reduced prompt example in Figure~\ref{fig:prompt_template}.

\paragraph{Model Input.} The model receives a prompt containing the public announcement, the agents' observations, and the history of the interaction up to the queried round as a single, non-interactive prompt. 

\paragraph{Model Output.} The model is asked regarding its own status (e.g., ``Is your forehead muddy?''). It must respond with exactly one of \textit{``Yes''}, \textit{``No''}, or \textit{``I don't know''}. In each round, exactly one response is correct based on the agent's current knowledge. 

To verify the model's response, we develop a symbolic solver grounded in Kripke possible worlds semantics, which is the standard framework for modeling knowledge in public-announcement settings~\cite{kripke1959completeness}.\footnote{See Appendix~\ref{app:solver_logic} for details and pseudocode.} In the symmetric inference setting, the label follows a closed-form formula to avoid unnecessary computation, whereas in the asymmetric inference setting, it is computed by an explicit simulation that iteratively eliminates possible worlds. This deterministic algorithm ensures correctness by construction.

\subsection{Benchmark}
To systematically isolate pattern-based from epistemic reasoning, our benchmark evaluates models along the two orthogonal dimensions in Figure~\ref{fig:benchmark}: narrative familiarity and inference complexity.

\paragraph{The Narrative Dimension (Horizontal).} This dimension isolates the impact of a puzzle's 
surface form across seven distinct narratives, as illustrated in the puzzle icons in Figure~\ref{fig:benchmark}. The classic \textit{Muddy Children}, \textit{Wise Men}, and \textit{Blue-Eyed Islanders} puzzles are canonical, appearing widely in textbooks and academic literature, and are therefore likely present in the pretraining data of LLMs. By contrast, the novel \textit{Safety Inspection}, \textit{Health Screening}, \textit{Olympic Games}, and \textit{Singing Contest} are puzzle narratives that we generate for this benchmark, differing substantially from the classics in setting, semantics, characters, objects, and surface vocabulary. We cross-reference these novel puzzles against known epistemic puzzle formulations and their key-words, and to the best of our knowledge no epistemic puzzle with these narratives exists in published or online sources.\footnote{See Appendix~\ref{app:prompt_examples} for prompt examples across all seven narratives and both inference types.}

Within each row of the matrix in Figure~\ref{fig:benchmark}, the underlying logical structure is preserved; announced bounds, observation rules, and interaction history remain fixed, 
while only the narrative changes. This structural parity lets us test whether models 
rely on pattern-based familiarity rather than reasoning over the provided epistemic 
structure.

\paragraph{The Inference Dimension (Vertical).} 
This dimension isolates reasoning difficulty. Moving down a column preserves the narrative framing while changing the underlying logic: standard symmetric setups, where agents observe everyone but themselves, are replaced by more complex asymmetric scenarios.
Inspired by \citet{sileo2023}, we formalize asymmetry using an \textit{observation matrix} $O \in \{0,1\}^{n \times n}$, where $O_{i,j}=1$ if and only if agent $i$ observes agent $j$'s status. Entries are sampled i.i.d.\ from $\{0,1\}$, yielding fragmented partial views that models must track.

\section{Experiments}
We evaluate LLMs on our benchmark dimensions.



\subsection{Experimental setup}

\paragraph{Models.}
We evaluate nine models from various families, including ChatGPT~\cite{chatgpt}, Gemini~\cite{gemini2023family}, Claude ~\cite{joshi2026architectural}, Mistral~\cite{jiang2023mistral7b}, Qwen~\cite{qwen3technicalreport} and OLMo~\cite{olmo}. We use the default temperature of 1 for the GPT-5 models due to API constraints and a temperature of 0 for all other models.

\paragraph{Dataset Generation.} We generate 100 instances per narrative across the seven narratives and both inference types, yielding 1,400 unique puzzles under a shared prompt structure. In the symmetric setup, the label distribution is 55 Yes and 45 No; in the asymmetric setup, it is 60 Yes and 40 No. Prompts are constructed using templates that map a shared  logical structure to narrative-specific vocabulary. By generating all narrative variants of a given instance from identical parameters, we guarantee by construction that they differ only in surface phrasing while maintaining logical equivalence.


\paragraph{Problem Instance Parameters.}
Our experiments use a fixed number of 10 agents, where the LLM is consistently agent 0. For symmetric inference, we cover all valid combinations of marked agents, lower bound values, and reasoning rounds of inference. The diagonal of the observation matrices is 0 ($O_{i,i}=0$) and the rest is filled with 1 ($O_{i,j}=1, i\neq j$). For asymmetric inference, we use a lower bound of 8, a reasoning round of 3 in all puzzles, randomly sampled observation matrices where the queried agent $0$ does not observe itself ($O_{0,0}=0$), and marked agents configured to allow inference in round 3. This design limits varying parameters to keep the comparison focused, as our preliminary experiments show it to be sufficiently challenging while still tractable.

Models are queried exclusively at the pivotal round, where an agent first accumulates sufficient information for deducing their own status. Since the prompt provides the full interaction history, including the agent's own past answers, a model can answer correctly in pre- and post-pivotal rounds simply from the prompt. At the pivotal round, however, the correct answer shifts from ``I don't know'' to a definitive ``Yes'' or ``No'' that cannot be read off the history, isolating the model's reasoning from its reliance on superficial cues.

\paragraph{Post-Process.} We evaluate responses by extracting only the first non-empty line, reflecting our prompt instructions. The output is mapped to exactly one of ``Yes'', ``No'', or ``I don't know'', allowing only minor formatting flexibility for whitespace, trailing periods, and apostrophe types, which proved sufficient for all evaluated models.


\begin{figure*}[!t]
    \centering
    \includegraphics[width=\textwidth]{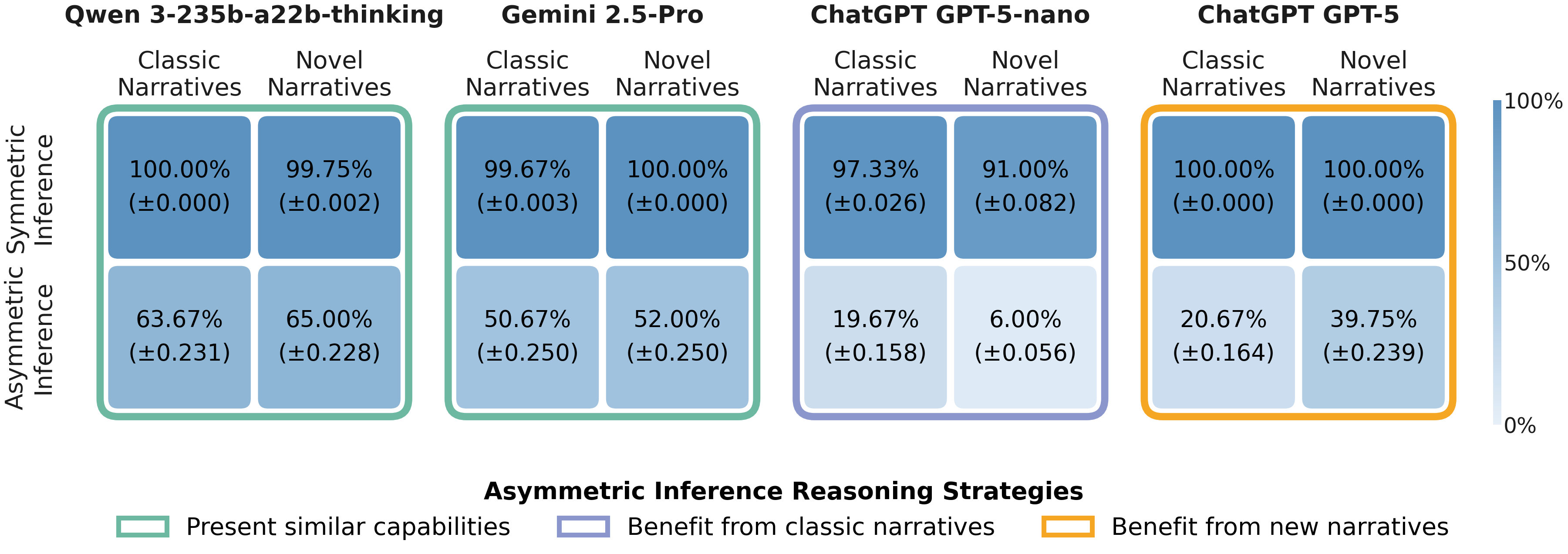}
    \caption{\textbf{Accuracy and variance of evaluated models across the two-dimensional benchmark.} While performance uniformly drops under asymmetric inference, models exhibit three distinct behavioral profiles based on narrative shifts in the asymmetric case: stable generalization (green), anchoring to familiar classic setups (purple), and improved performance when novel narratives strip away learned biases (orange). In each quadrant near-zero variance ($\pm$) indicates uniform success or consistent failure; high variance reflects fragmented behavior.}
    \label{fig:success_rates}
\end{figure*}

\subsection{Results}
Since some models fail to outperform majority vote on the \textit{Muddy Children} puzzle under symmetric inference and mostly reuse the same response (e.g., Olmo 3-7B-Instruct answers ``No'' in 99\% of instances), we exclude them from the main analysis.\footnote{See Appendix~\ref{sec:taskDef} for additional results.}
Several observations can be drawn from our results.

\paragraph{Models demonstrate a strong ability to generalize from classic to novel narratives within the symmetric setup, indicating strong pattern-based reasoning capabilities.}
Figure~\ref{fig:success_rates} shows that when moving from classic to novel narratives under symmetric inference, most of the models successfully solve the new instances. This robustness to narrative modifications contrasts with prior findings, such as those of ~\citet{jiang2024} and ~\citet{mirzadeh2025gsm}, which observed a reversed trend where token perturbations significantly degraded model performance.

\paragraph{In the asymmetric setting, performance drops sharply, highlighting both inference complexity and a reliance on pattern-based reasoning.}
When the task shifts to asymmetric setups that require tracking possible worlds and can't be solved with the closed-form formula applicable in the symmetric setup, accuracy declines sharply across all models. This shift is also reflected in the reported variance across the quadrants of Figure~\ref{fig:success_rates}: symmetric setups yield near-zero variance (models are consistently correct or incorrect), whereas asymmetric setups show higher variance, indicating mixed success across puzzles. While this is inherently a more complex task, both inference setups rely on the same foundational Kripke possible worlds semantics~\cite{kripke1959completeness}, and solving from first principles means the same procedure in both. The difficulty arises not just from structural complexity, but from a lack of familiarity, given that novel puzzle formulations are mostly concentrated in the symmetric setting where the process collapses to a closed-form formula. This suggests that models have limited epistemic deduction abilities.


\paragraph{Under asymmetric inference, models differ in sensitivity to narrative changes.} Some models, such as Qwen 3-235b-a22b-thinking and Gemini 2.5-Pro, show consistently low performance across narratives, suggesting that the main bottleneck is the asymmetric epistemic structure itself. ChatGPT GPT-5-nano, on the other hand, performs better on classic narratives, indicating a stronger reliance on recognizable templates. In contrast, ChatGPT GPT-5 performs better on novel narratives, suggesting that classic narratives may trigger misleading associations rather than support the required inference.\footnote{All comparisons are statistically significant ($p < 10^{-4}$).}

\section{Conclusion}
We revisit the evaluation of LLMs on epistemic puzzles, arguing that the contrast between epistemic reasoning and memorization is too coarse for newer models. We show that strong performance seems to come from reducing to familiar problem structures rather than performing epistemic reasoning.
Future work can examine model performance within interactive settings, as well as testing models in a tool-augmented condition, where state tracking is offloaded to external memory.

\section{Limitations}
Our benchmark focuses on DEL-style hidden-status puzzles, which capture central mechanisms of epistemic reasoning: partial observation, higher-order knowledge, public updates, and reasoning from others' uncertainty. However, they do not cover the full range of epistemic phenomena, such as private announcements, belief revision, deception, strategic communication, or planning under uncertainty.

\section*{Acknowledgments}
We thank the anonymous reviewers for their helpful feedback and comments.
This work was partially supported by research grant
no.~7256 from the Israeli Ministry of Science and
Technology, and  by the Israel Science Foundation grant no.~2355/24.




\bibliography{custom}
\clearpage

\appendix

\section{Appendix - Prompt Examples, Solver Logic and Additional Results}\label{app:prompt_structure}
\subsection{Prompt examples}\label{app:prompt_examples}
\begin{figure*}[t]
    \centering
    \includegraphics[width=0.98\textwidth]{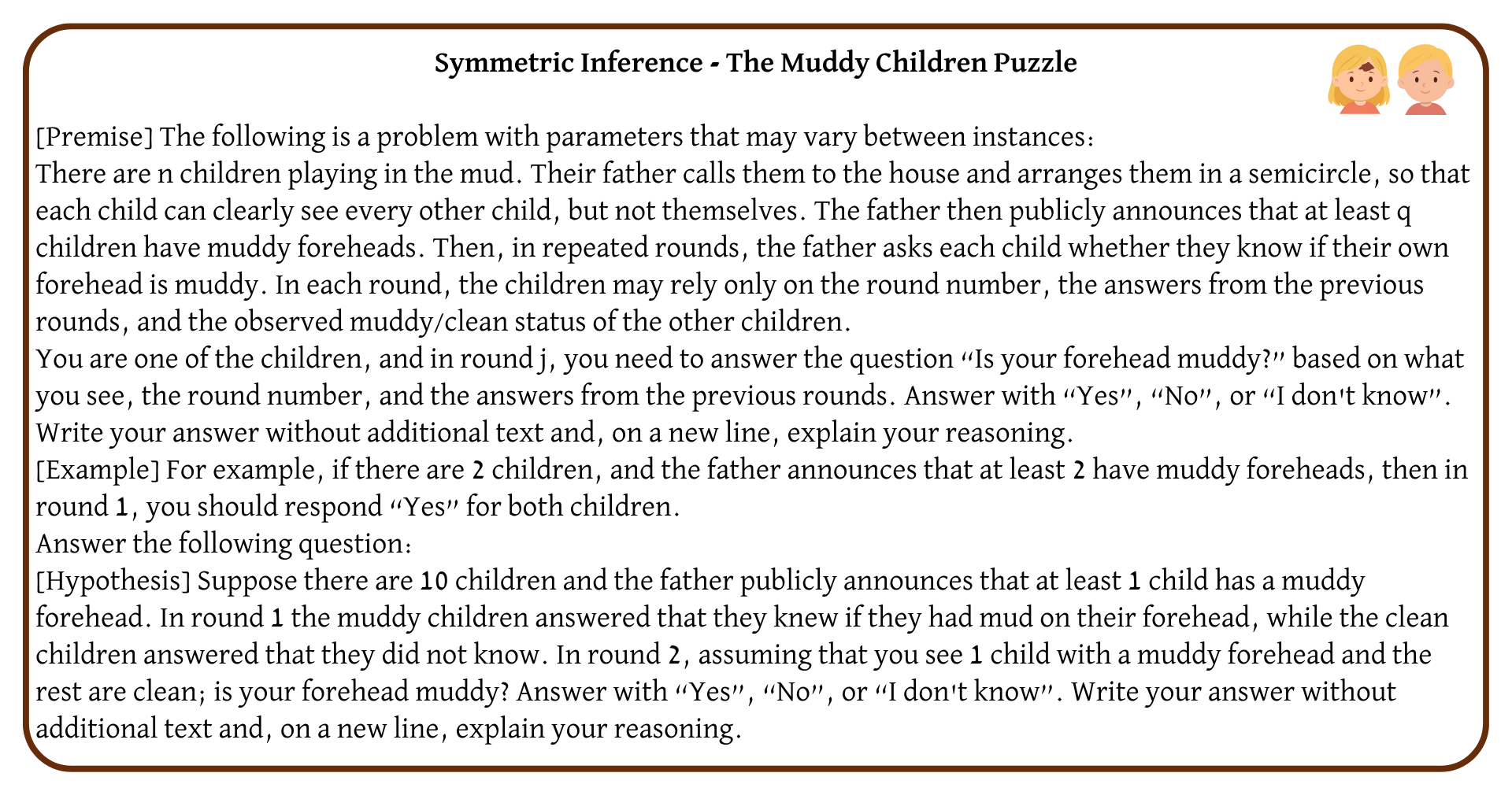}
    \caption{Prompt example - symmetric inference, the \textit{Muddy Children} puzzle.}
    \label{fig:25}
\end{figure*}

\begin{figure*}[t]
    \centering
    \includegraphics[width=0.98\textwidth]{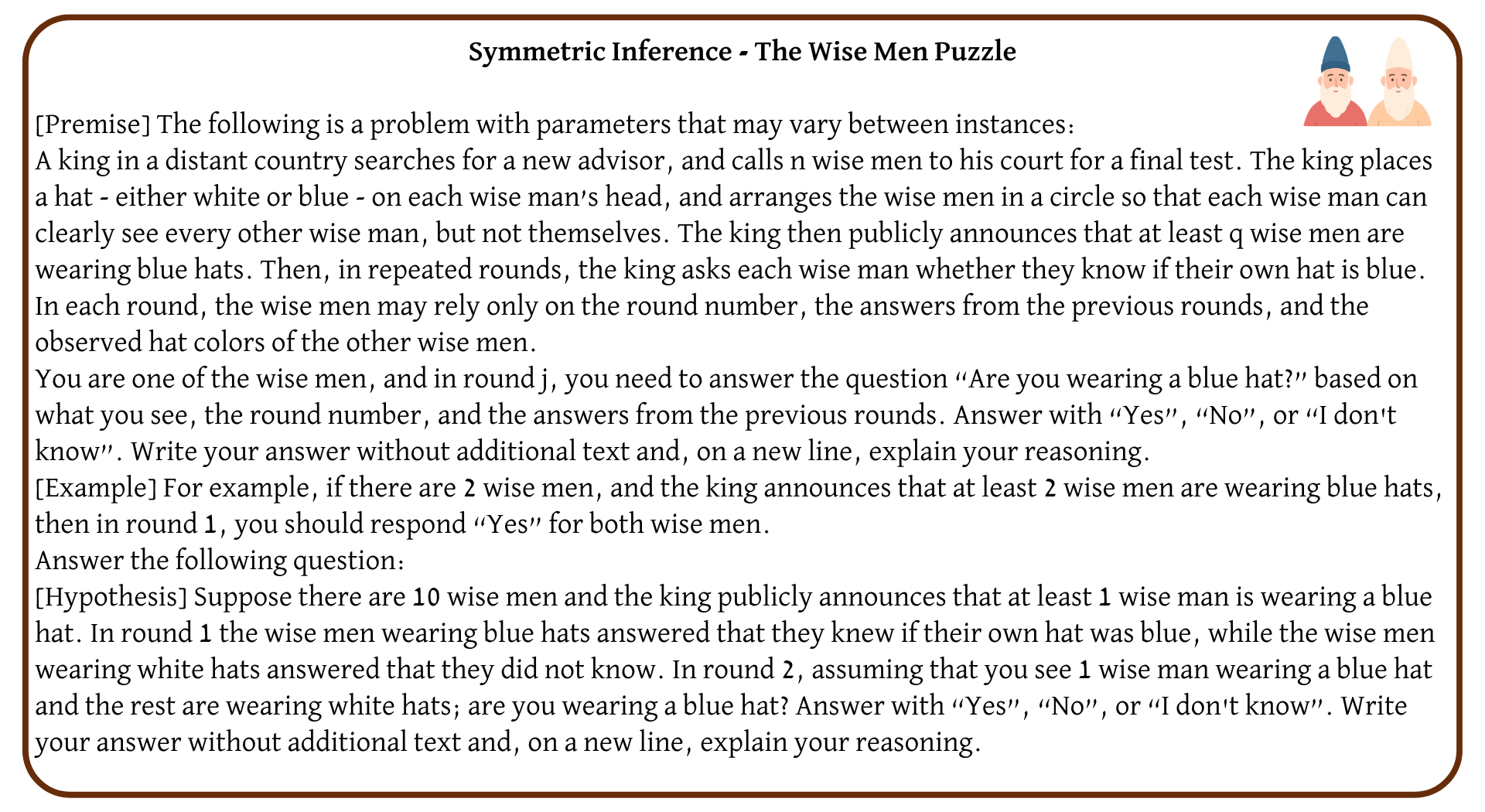}
    \caption{Prompt example - symmetric inference, the \textit{Wise Men} puzzle.}
    \label{fig:26}
\end{figure*}

\begin{figure*}[t]
    \centering
    \includegraphics[width=0.98\textwidth]{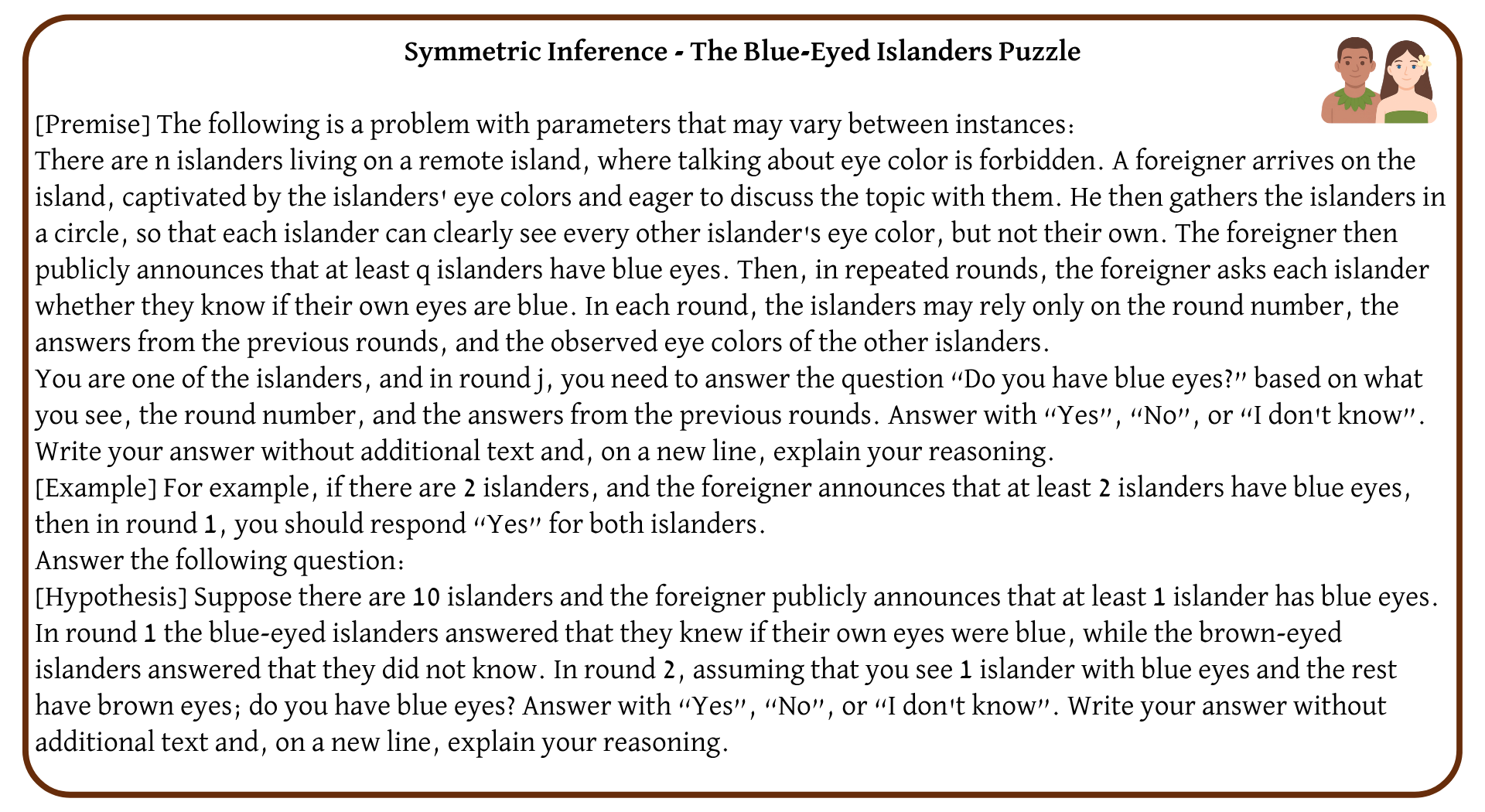}
    \caption{Prompt example - symmetric inference, the \textit{Blue-Eyed Islanders} puzzle.}
    \label{fig:27}
\end{figure*}

\begin{figure*}[t]
    \centering
    \includegraphics[width=0.98\textwidth]{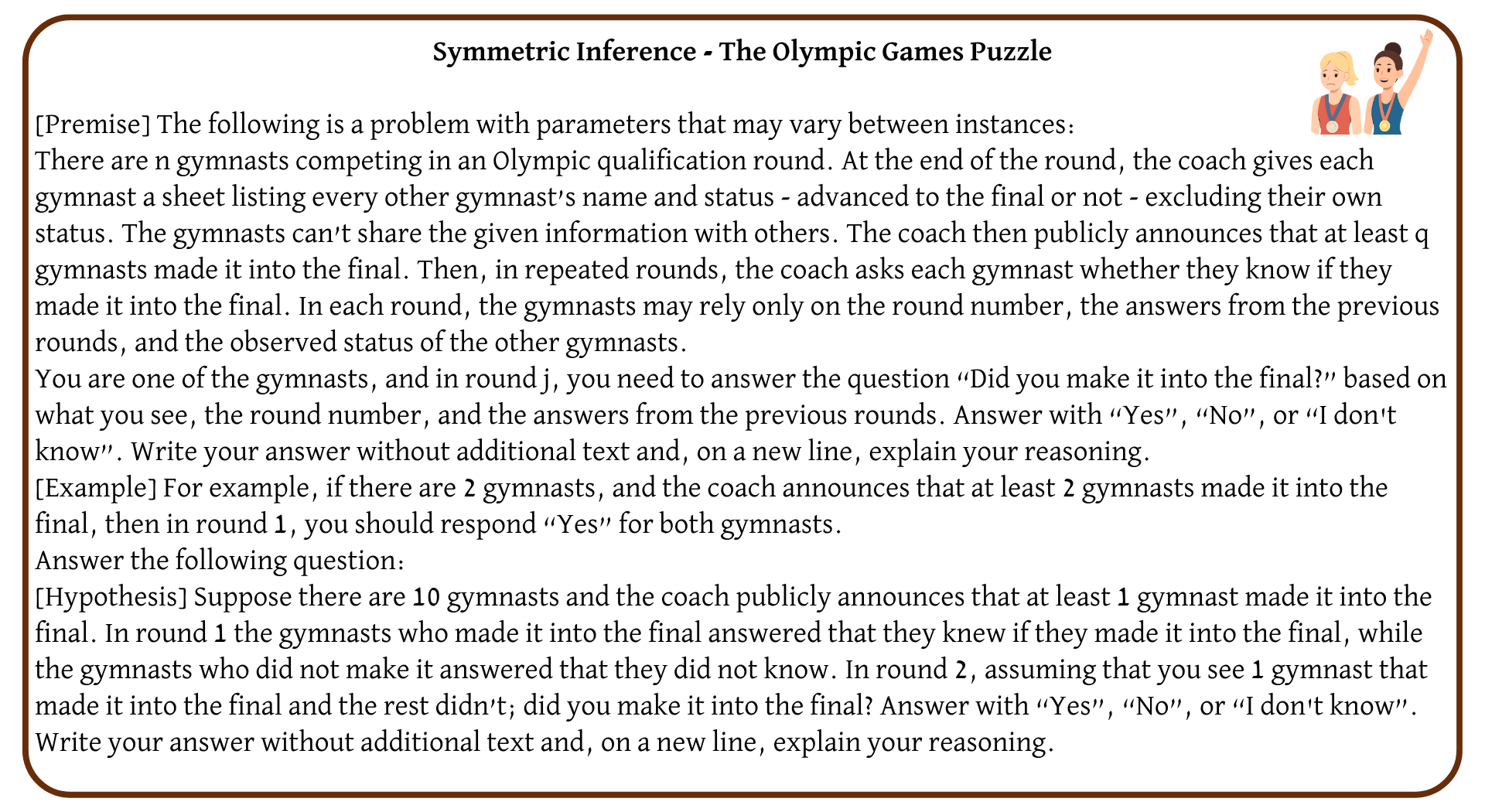}
    \caption{Prompt example - symmetric inference, the \textit{Olympic Games} puzzle.}
    \label{fig:28}
\end{figure*}

\begin{figure*}[t]
    \centering
    \includegraphics[width=0.98\textwidth]{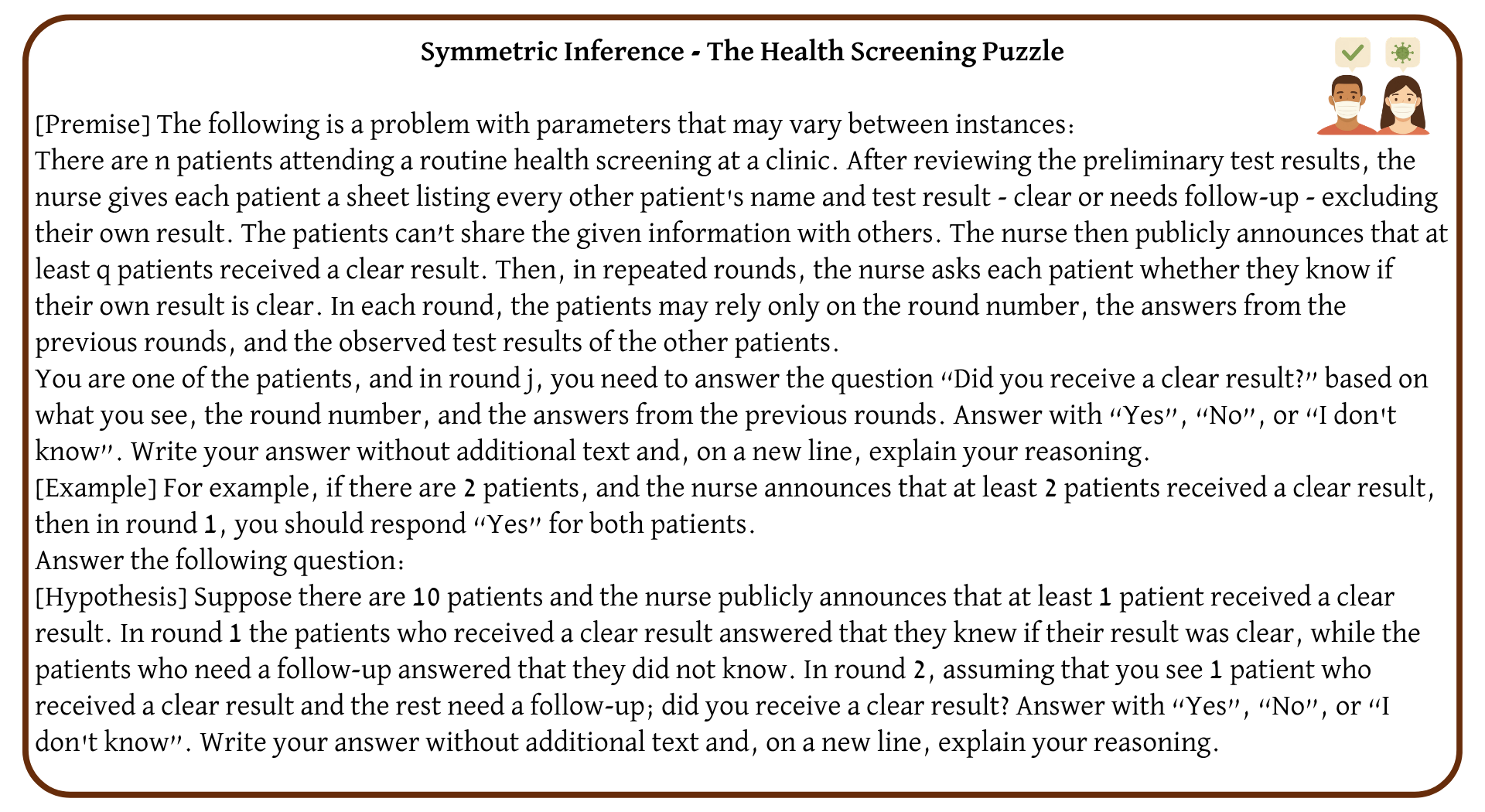}
    \caption{Prompt example - symmetric inference, the \textit{Health Screening} puzzle.}
    \label{fig:29}
\end{figure*}

\begin{figure*}[t]
    \centering
    \includegraphics[width=0.98\textwidth]{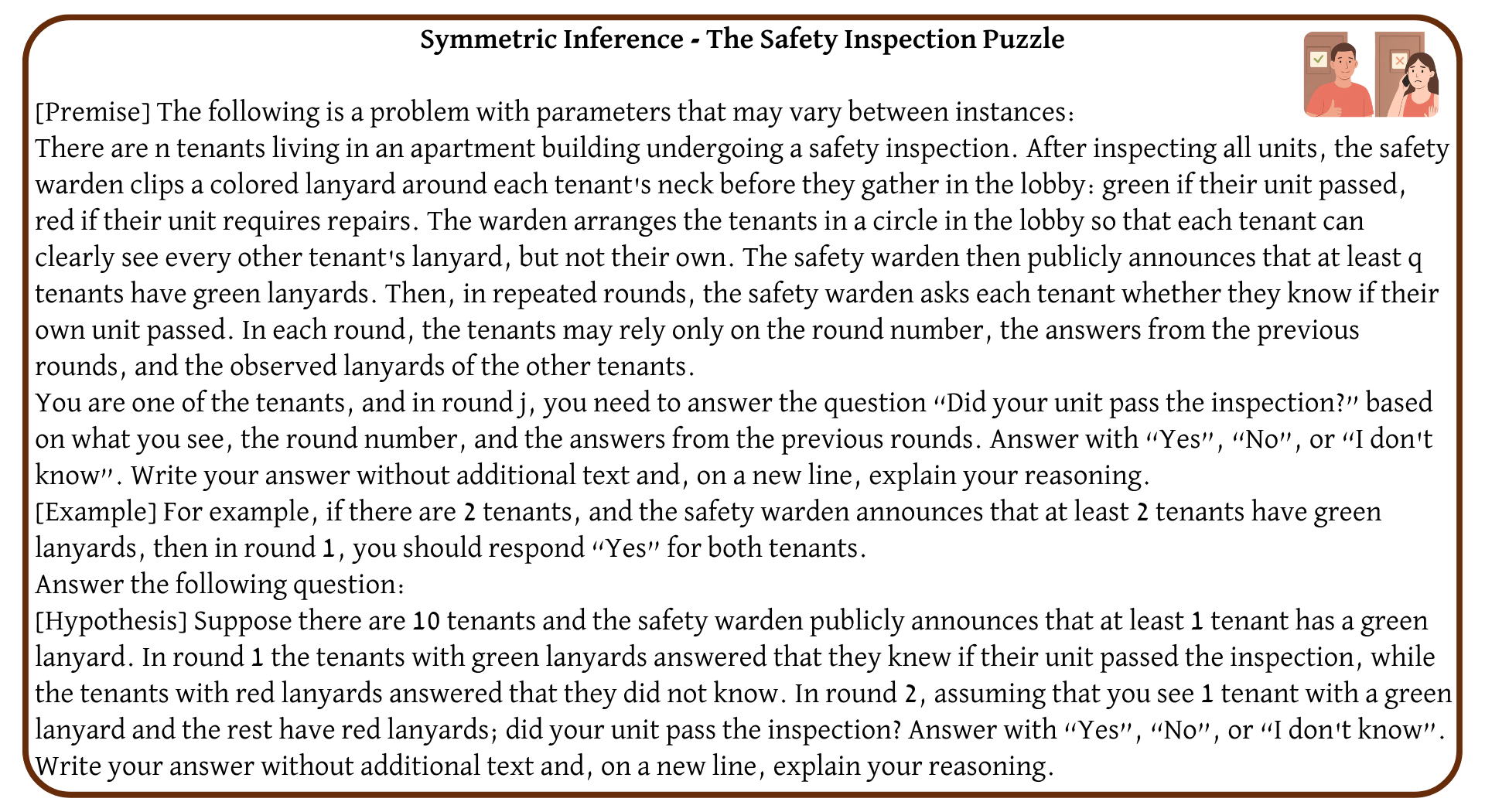}
    \caption{Prompt example - symmetric inference, the \textit{Safety Inspection} puzzle.}
    \label{fig:30}
\end{figure*}

\begin{figure*}[t]
    \centering
    \includegraphics[width=0.98\textwidth]{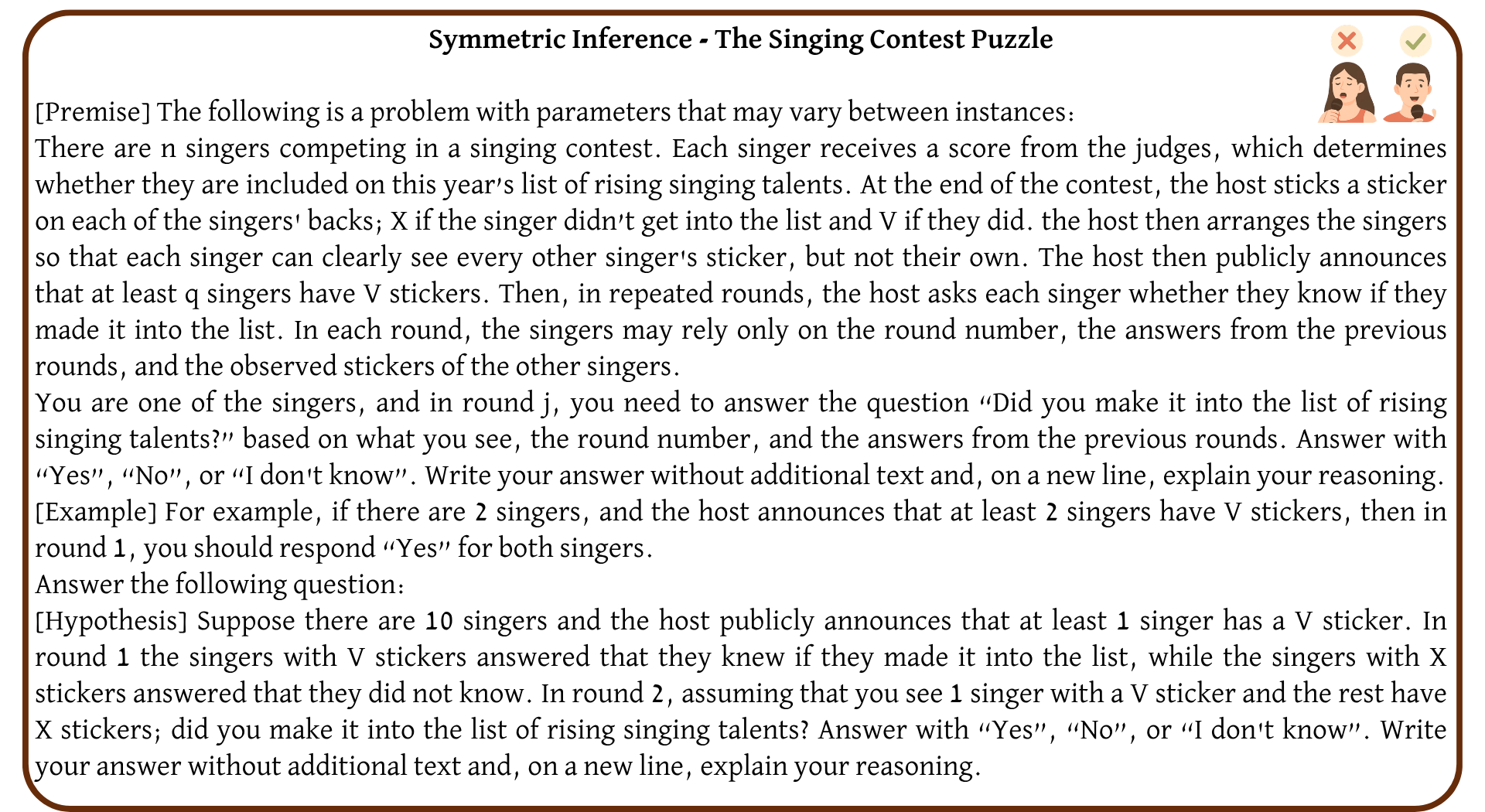}
    \caption{Prompt example - symmetric inference, the \textit{Singing Contest} puzzle.}
    \label{fig:31}
\end{figure*}

\begin{figure*}[t]
    \centering
    \includegraphics[width=0.98\textwidth]{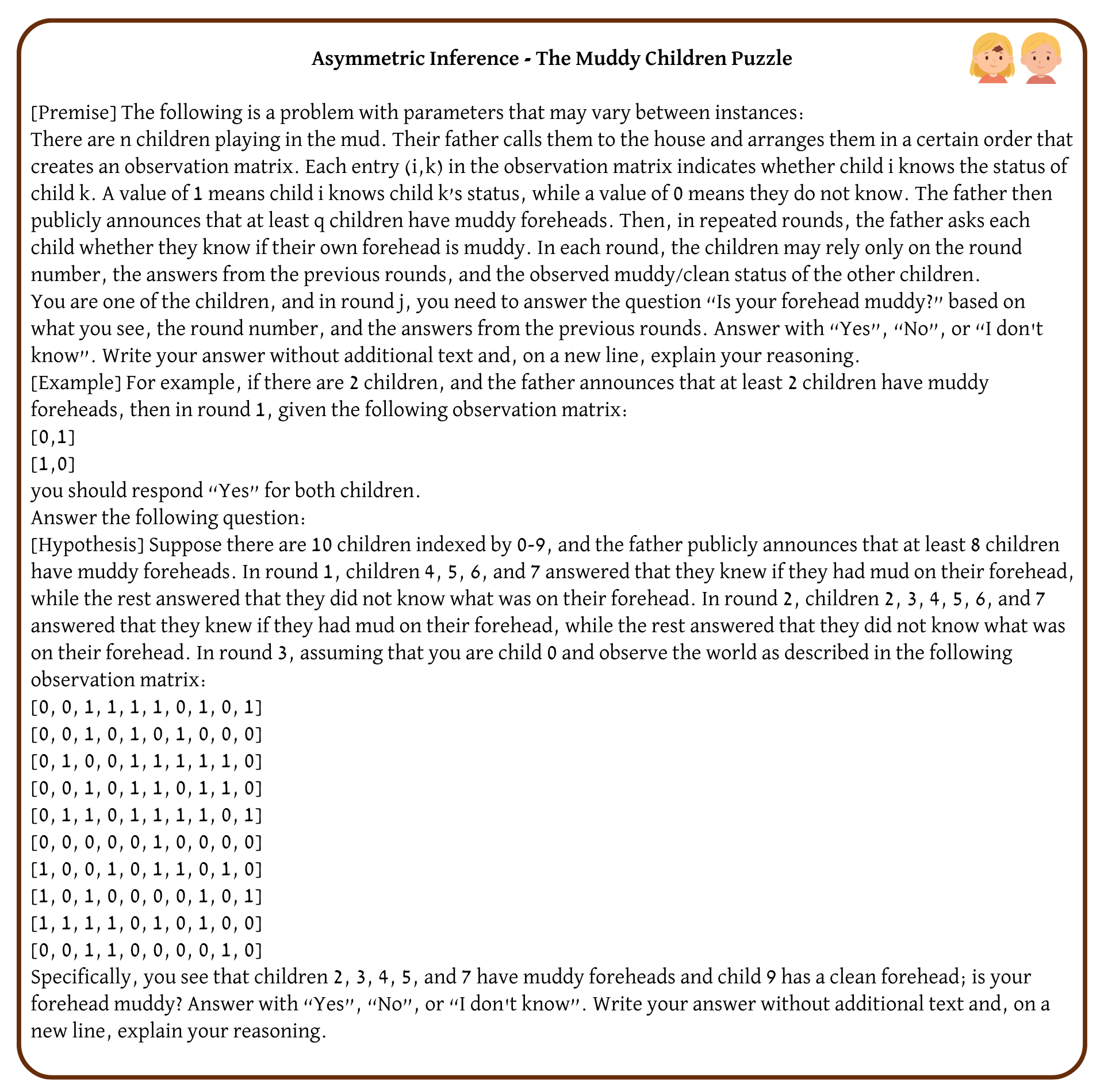}
    \caption{Prompt example - asymmetric inference, the \textit{Muddy Children} puzzle.}
    \label{fig:32}
\end{figure*}

\begin{figure*}[t]
    \centering
    \includegraphics[width=0.98\textwidth]{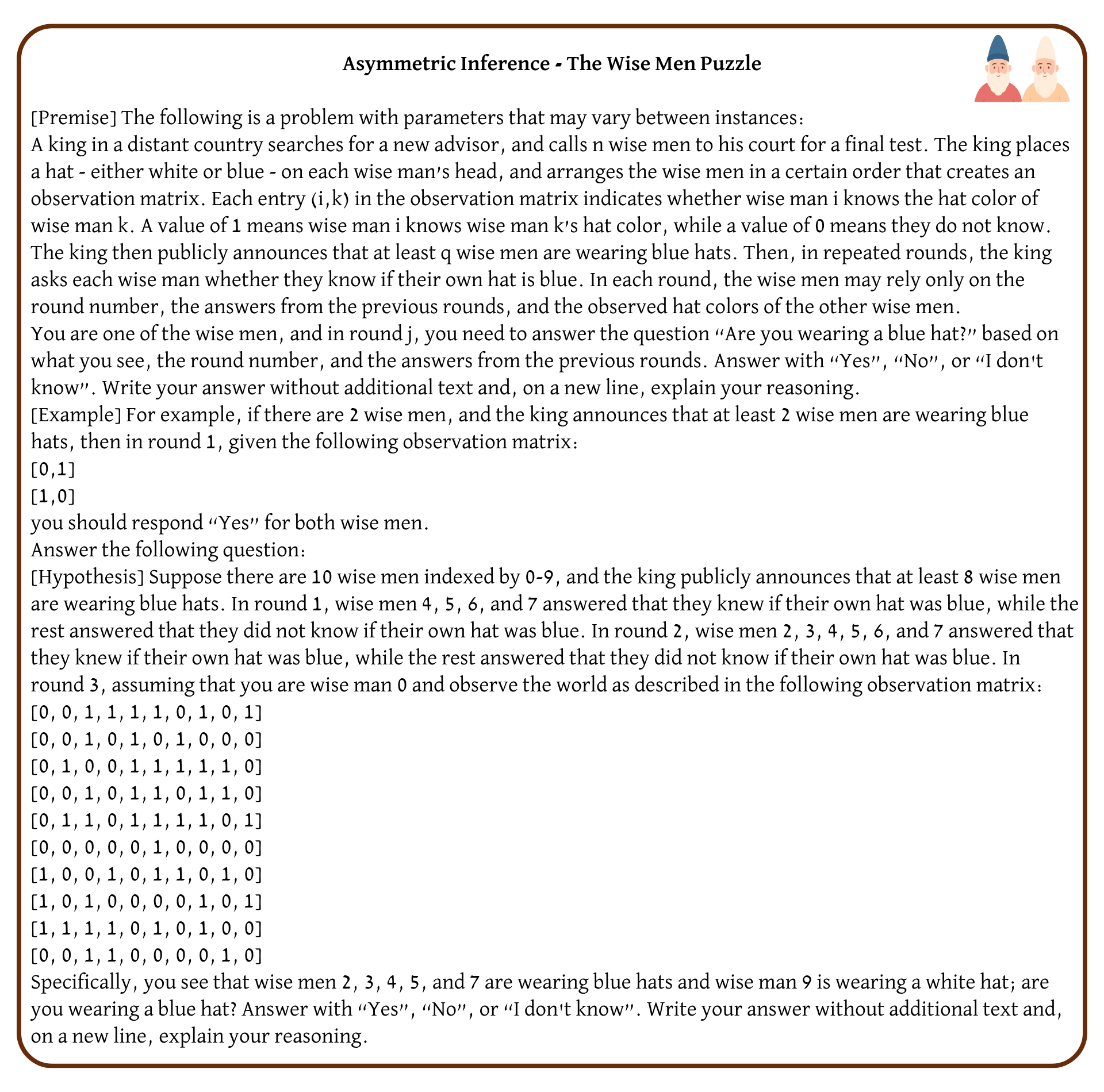}
    \caption{Prompt example - asymmetric inference, the \textit{Wise Men} puzzle.}
    \label{fig:33}
\end{figure*}

\begin{figure*}[t]
    \centering
    \includegraphics[width=0.98\textwidth]{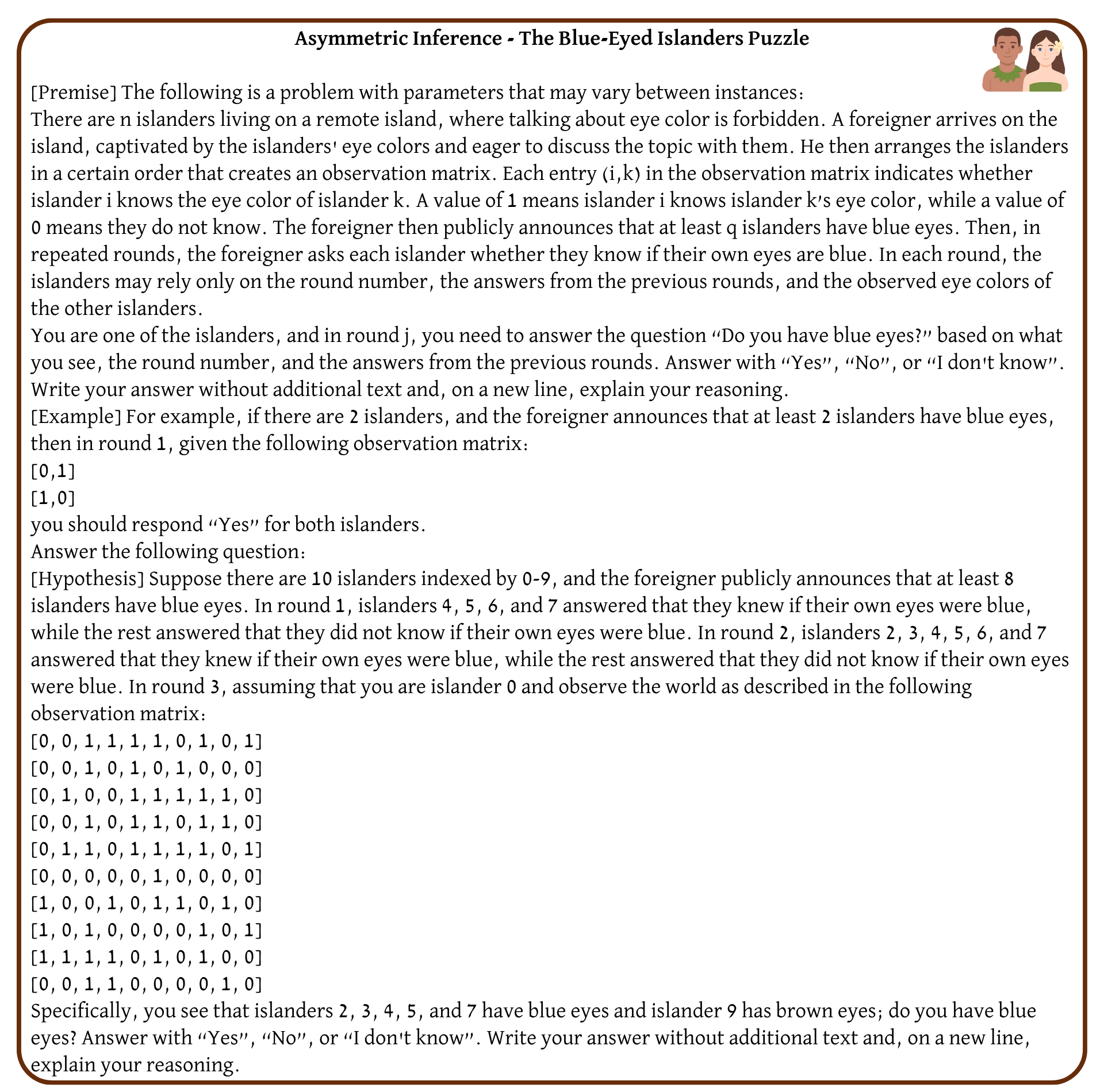}
    \caption{Prompt example - asymmetric inference, the \textit{Blue-Eyed Islanders} puzzle.}
    \label{fig:34}
\end{figure*}

\begin{figure*}[t]
    \centering
    \includegraphics[width=0.98\textwidth]{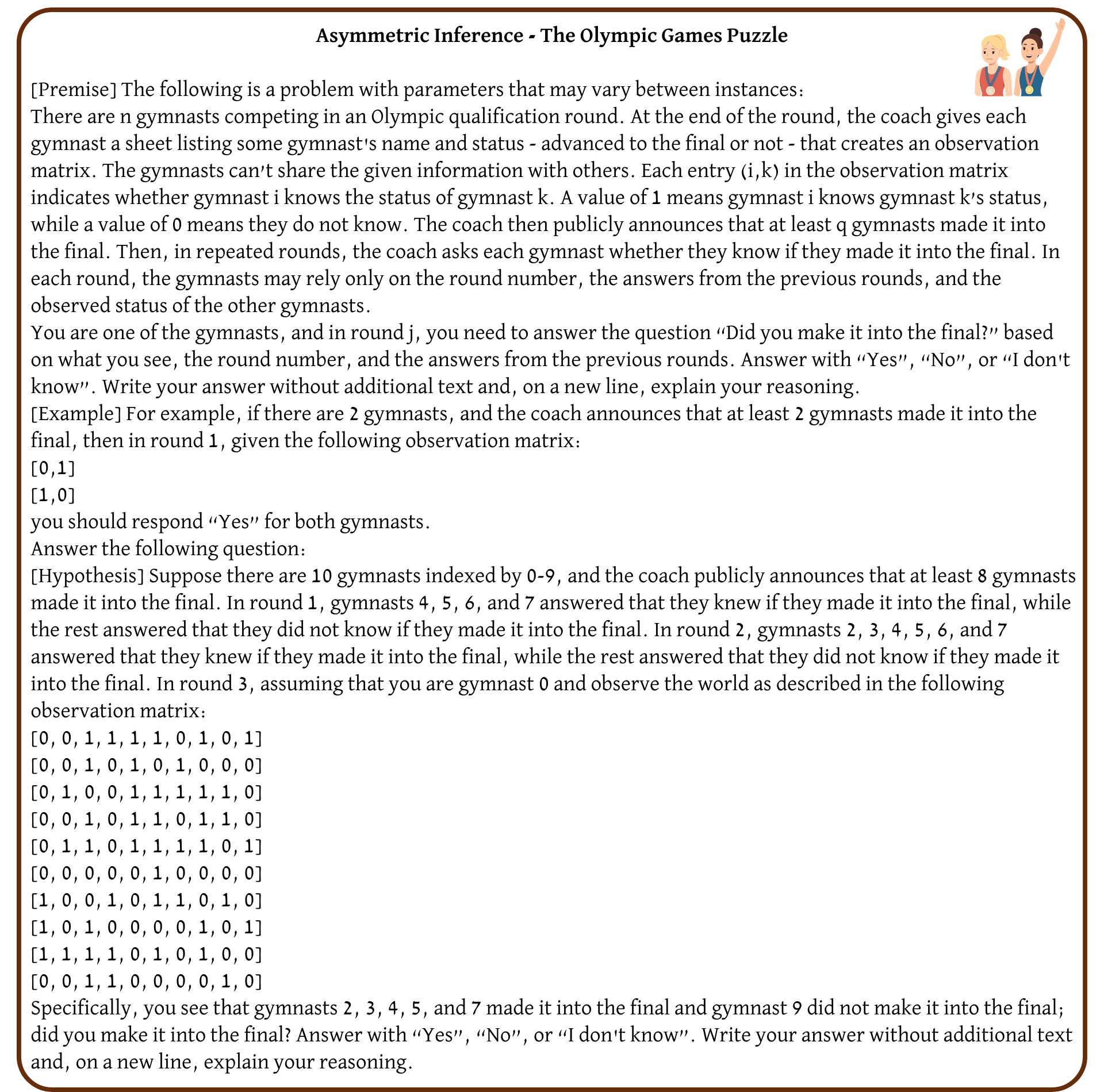}
    \caption{Prompt example - asymmetric inference, the \textit{Olympic Games} puzzle.}
    \label{fig:35}
\end{figure*}

\begin{figure*}[t]
    \centering
    \includegraphics[width=0.98\textwidth]{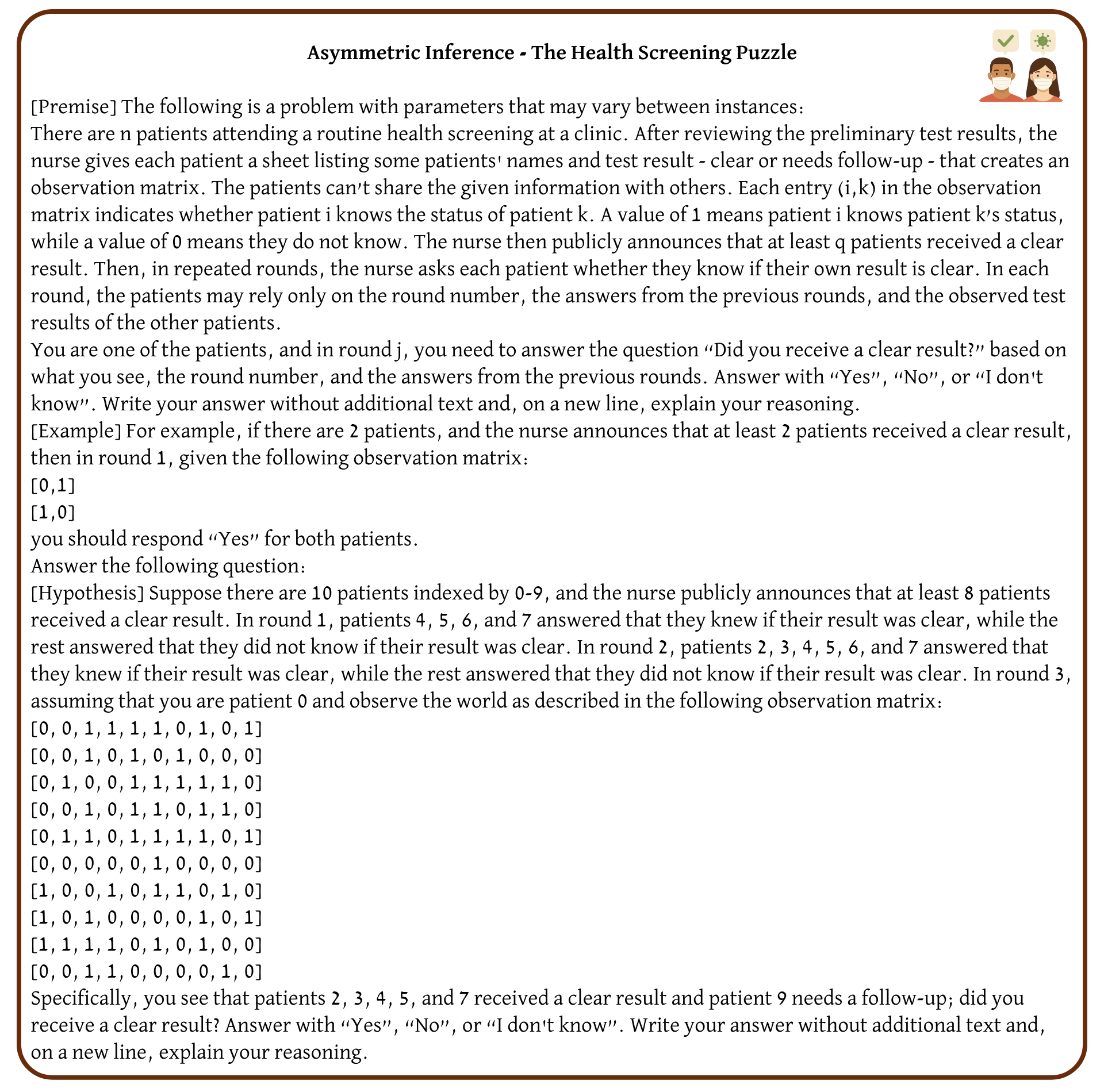}
    \caption{Prompt example - asymmetric inference, the \textit{Health Screening} puzzle.}
    \label{fig:36}
\end{figure*}

\begin{figure*}[t]
    \centering
    \includegraphics[width=0.98\textwidth]{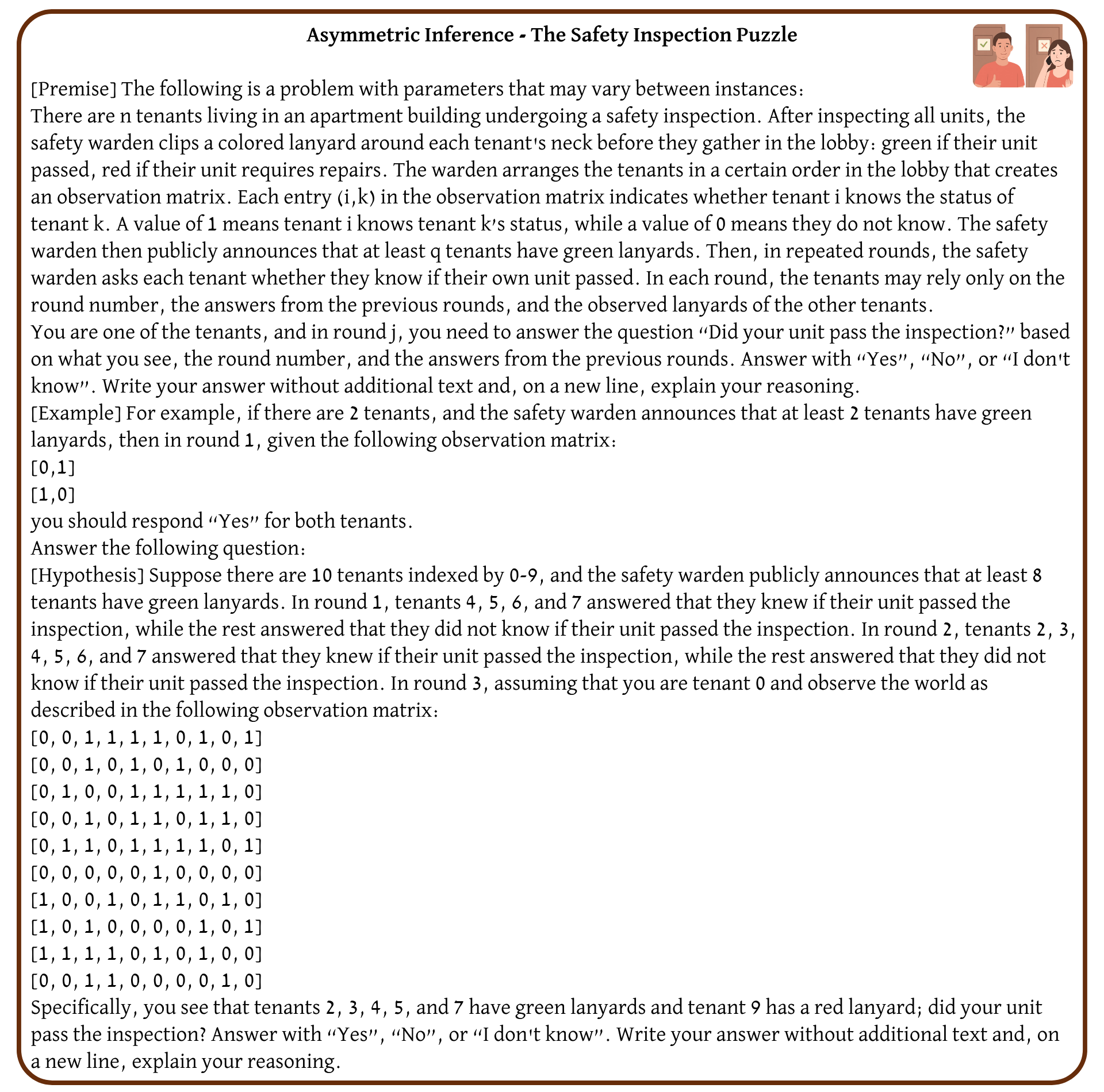}
    \caption{Prompt example - asymmetric inference, the \textit{Safety Inspection} puzzle.}
    \label{fig:37}
\end{figure*}

\begin{figure*}[t]
    \centering
    \includegraphics[width=0.98\textwidth]{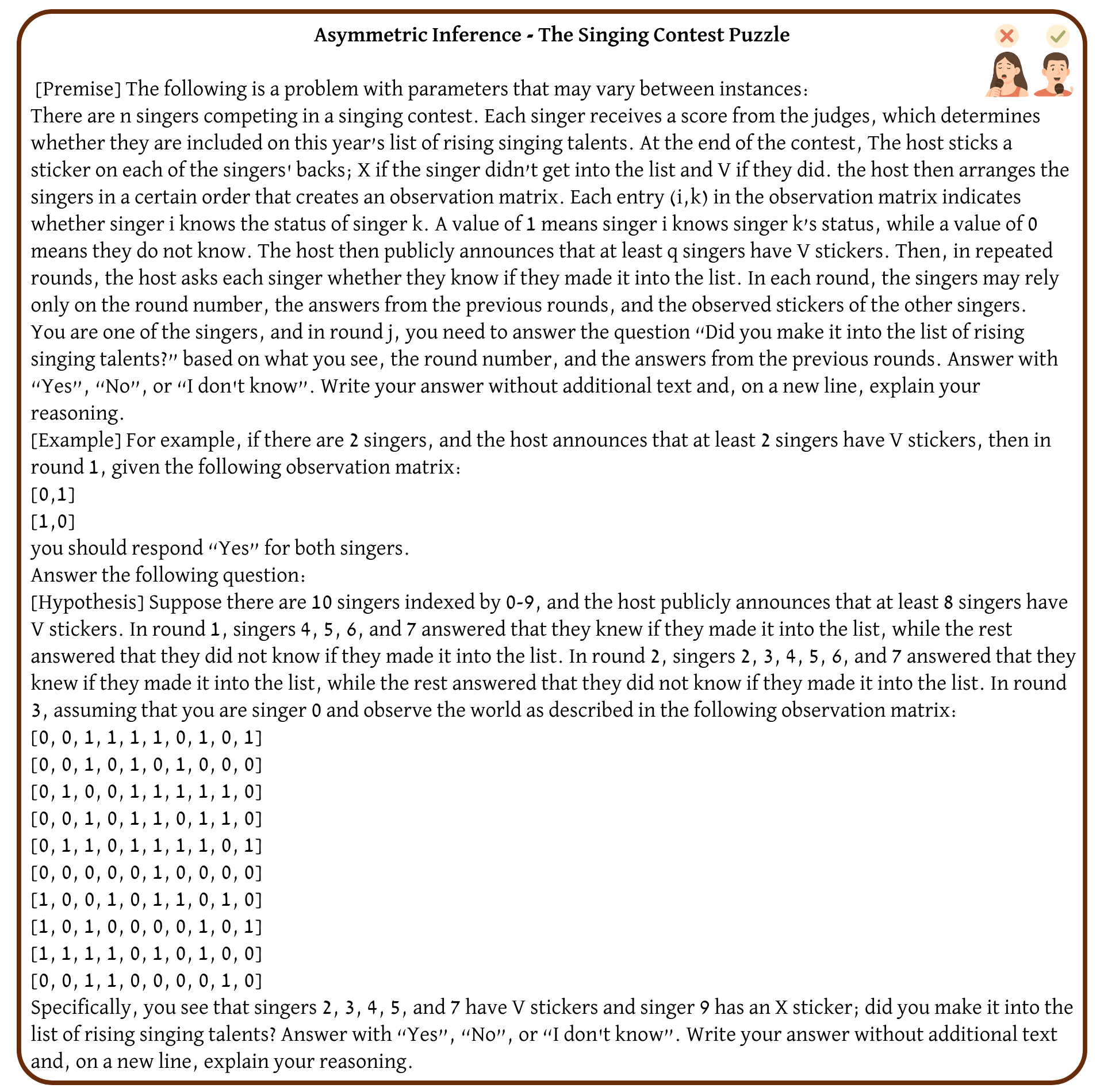}
    \caption{Prompt example - asymmetric inference, the \textit{Singing Contest} puzzle.}
    \label{fig:38}
\end{figure*}

Figures~\ref{fig:25}-\ref{fig:31}
show prompt examples for the 7 puzzle narratives under the symmetric inference, all of the same parameters with the narrative modified.
Figures~\ref{fig:32}-\ref{fig:38} show prompt examples for the 7 puzzle narratives under the asymmetric inference, all of the same parameters with the narrative modified.

\begin{algorithm}[t]
\caption{\textsc{GetSolverLabel}}
\label{alg:solver-label}
{\small\textit{Determines whether agent $i$ knows their status at round $j$, handling both the symmetric (closed-form) and asymmetric (pre-computed history) cases.}}
\begin{algorithmic}[1]
\Require $n$: total agents;\; $k$: positive agents;\; $j$: query round;\;
         $q$: boundary value;\; $\tau \in \{\textsc{lower},\textsc{upper}\}$: bound type;\;
         $i$: queried agent;\; $\mu \in \{0,1\}$: agent $i$'s true status;\;
         $\mathit{rand}$: \textsc{true} for asymmetric inference;\;
         $\mathcal{H}$: knowledge history (only used in the asymmetric case, to avoid excessive computations.)
\Ensure  $1$ if agent $i$ knows their status at round $j$, else $0$

\If{$\mathit{rand}$}
    \State \Return $i \in \mathcal{H}[j]$
    \Comment{pre-computed by Algorithm~\ref{alg:knowledge-history}}
\EndIf
\If{$(\tau {=} \textsc{lower} \wedge q {=} 0) \vee (\tau {=} \textsc{upper} \wedge q {=} n)$}
    \State \Return $0$ \Comment{uninformative bound}
\EndIf

\Comment{If in the symmetric case, compute with closed formula}
\State $s \gets |k - q| + 1$
\Comment{first round at which some agent gains knowledge}
\If{$(\tau {=} \textsc{lower} \wedge \mu {=} 1) \vee (\tau {=} \textsc{upper} \wedge \mu {=} 0)$}
    \State \Return $j \ge s$
    \Comment{positives under lower bound know first, negatives under upper bound know first}
\Else
    \State \Return $j \ge s+1$
    \Comment{the other group knows one round later}
\EndIf
\end{algorithmic}
\end{algorithm}

\begin{algorithm}[!t]
\caption{Iterated Possible-Worlds Elimination}
\label{alg:knowledge-history}
{\small\textit{Simulates iterative reasoning  grounded in \textbf{Kripke possible worlds semantics~\cite{kripke1959completeness}}, computing for each round which agents know their own status.}}
\begin{algorithmic}[1]
\Statex \textbf{procedure} \textsc{GetKnowledgeHistory}$(n, O, w^{*}, q, \tau)$
\Require $n$: total agents;\; $O \in \{0,1\}^{n \times n}$: observation matrix;\;
         $w^{*} \in \{0,1\}^{n}$: true world;\;
         $q$: boundary value;\; $\tau$: bound type
\Ensure  $\mathcal{H}$: list of knower sets for each round from 1 to $n{+}1$: $\mathcal{H}[1], \ldots, \mathcal{H}[n{+}1]$

\State $\mathcal{W} \gets \{w \in \{0,1\}^{n} : w \text{ consistent with } (\tau, q)\}$
\State $\mathcal{H} \gets [\,]$
\For{$t = 1$ \textbf{ to } $n{+}1$}
    \State $K_{t} \gets \mathit{FindKnowers}(n, O, w^{*}, \mathcal{W})$
    \State Append $K_{t}$ to $\mathcal{H}$
    \If{$|\mathcal{W}| \le 1$}
        \State Fill $\mathcal{H}$'s remaining slots with $K_{t}$; \textbf{break}
    \EndIf
    \State $\mathcal{W}' \gets \{w \in \mathcal{W} : \mathit{FindKnowers}(n,O,w,\mathcal{W})=K_{t}\}$
    \If{$\mathcal{W}' = \mathcal{W}$}
        \Comment{steady state}
        \State Fill $\mathcal{H}$'s remaining slots with $K_{t}$; \textbf{break}
    \EndIf
    \State $\mathcal{W} \gets \mathcal{W}'$
\EndFor
\State \Return $\mathcal{H}$

\Statex

\Statex \textbf{procedure} \textsc{FindKnowers}$(n, O, w^{*}, \mathcal{W})$
    \Require $n$: total agents;\; $O \in \{0,1\}^{n \times n}$: observation matrix;\;
         $w^{*} \in \{0,1\}^{n}$: true world; $\mathcal{W}$: current candidate worlds
    \Ensure  $K$: indices of agents that know their own status in the current world based on the revealed information.
    \State $K \gets [\,]$
    \For{$i = 0$ \textbf{ to } $n{-}1$}
        \State $V_{i} \gets \{i' : O_{i,i'} = 1\}$
        \Comment{agents observable by $i$}
        \State $\mathcal{W}_{i} \gets \{w \in \mathcal{W} : w[V_{i}] = w^{*}[V_{i}]\}$
        \Comment{worlds matching $i$'s view}
        \If{$\mathcal{W}_{i} \ne \emptyset \;\wedge\; |\{w[i] : w \in \mathcal{W}_{i}\}| = 1$}
            \State Append $i$ to $K$
        \EndIf
    \EndFor
    \State \Return $K$

\end{algorithmic}
\end{algorithm}

\subsection{Solver Logic}\label{app:solver_logic}
Our solver is grounded in Kripke possible worlds semantics~\cite{kripke1959completeness}, and assigns a ground-truth label to every problem instance, determining whether a queried agent knows their own status at a given reasoning round.
The solver handles two inference settings through a unified interface
(Algorithm~\ref{alg:solver-label}).

\paragraph{Symmetric inference.}
When all agents observe all others (i.e., the observation matrix $O$
has $O_{i,i'}=1$ for all $i \neq i'$), the epistemic state of every
agent can be derived analytically.
Given the public announcement bound $(\tau, q)$ and the true number
of positive agents $k$, the first round at which any agent gains
knowledge is $s = |k - q| + 1$.
Under a lower bound, positive agents ($\mu=1$) know at round $s$
and negative agents at round $s+1$; the roles are reversed under
an upper bound.
\textsc{ProduceSolverLabel} (Algorithm~\ref{alg:solver-label})
encodes this closed-form rule directly, returning $1$ if and only
if the query round $j$ meets or exceeds the agent's knowledge
threshold.
If the bound is uninformative (e.g., ``at least $0$'' or
``at most $n$''), no agent can ever know, and the label is $0$.

\paragraph{Asymmetric inference.}
When agents observations of each other are detailed through an arbitrary binary matrix
$O \in \{0,1\}^{n \times n}$, there is no closed form.
Algorithm~\ref{alg:knowledge-history} implements \emph{iterated
possible world elimination}, pre-computes a knowledge history $\mathcal{H}$ starting from the set $\mathcal{W}$ of all worlds consistent with
the public announcement. 

\begin{table*}[]
\centering
\caption{Accuracy (\%) of omitted models over the \textit{Muddy Children} puzzle with symmetric inference.}
\label{tab:model_accuracy}
\begin{tabular}{lc}
\toprule
\textbf{Model} & \textbf{Accuracy (\%)} \\
\midrule
ChatGPT GPT-4                   & 52\% \\
Gemini 2.5 Flash Lite           & 48\% \\
OLMo 3-7B-Instruct              & 46\% \\
Claude Opus-4.6              & 46\% \\
Mistral Small-24B-Instruct-2501 & 52\% \\
\bottomrule
\end{tabular}
\end{table*}

\subsection{Experimental Setup, Evaluated Models and Additional Results}
\paragraph{Problem Instance structure.}\label{sec:taskDef}
We define a problem instance using the tuple $(n,k,q,j,O)$. In this formulation, $n$ represents the total number of agents (e.g., the number of children in the \textit{Muddy Children} puzzle), and $k$ denotes the count of positive agents possessing the hidden status (e.g., being ``muddy''). The parameter $q$ represents the lower bound value in the public announcement (e.g., ``at least $q$ are muddy''), while $j$ specifies the target reasoning round at which the agent's knowledge is evaluated. Finally, $O \in \{0, 1\}^{n \times n}$ is an observation matrix where the entry $O_{i,i'}=1$ if agent $i$ observes the status of agent $i'$, and $O_{i,i'}=0$ otherwise. 

An example of this parameterization is illustrated in Figure \ref{fig:reasoning_rounds}. There, the number of agents is $n=2$, the number of agents possessing the hidden status (muddy) is $k=1$, the boundary is a lower bound of value $q=1$, and the observations are as in the matrix $O=\begin{pmatrix}0 & 1 \\ 1 & 0\end{pmatrix}$, meaning each child observes the other child, but not themselves.

Our experiments use a fixed number of agents $n=10$, where the LLM is always agent 0.

\paragraph{Pivotal Round Evaluation.} Models are queried exclusively at the pivotal round, where an agent first accumulates sufficient information for deducing their own status. Since the prompt provides the full interaction history, including the agent's own past answers, a model can answer correctly in pre- and post-pivotal rounds simply from the prompt. At the pivotal round, however, the correct answer shifts from ``I don't know'' to a definitive ``Yes'' or ``No'' that cannot be read off the history, isolating the model's reasoning from its reliance on superficial cues.

\paragraph{Symmetric Inference.} We cover all valid combinations of marked agents $k\in[0,10]$, lower bound value $q\in[0,k]$, and pivotal reasoning rounds of inference. The diagonal of the observation matrices is 0 ($O_{i,i}=0$) and the rest is filled with 1 ($O_{i,j}=1, i\neq j$).

\paragraph{Asymmetric Inference.} We use a lower bound $q=8$, a reasoning round of $j=3$ in all puzzles, and binary observation matrices sampled randomly from $\{0, 1\}^{n \times n}$, ensuring within the prompt that the queried agent does not observe itself ($O_{0,0}=0$). The marked agents are configured to allow inference in reasoning round 3. We restrict the asymmetric settings to round 3 since rounds 1 and 2 examine shallow inference (a direct read of the boundary condition, or a single elimination over the first round's declarations), whereas round 3 is the earliest round requiring nested reasoning about what agents conclude from one another. The bound $q=8$ follows from this restriction, since instances solvable exactly at round 3 concentrate around this value. Fixing these values limits the number of varying parameters and keeps the comparison focused on our two dimensions of interest, as our preliminary experiments show this specific design to be sufficiently challenging while still tractable.

\begin{table*}[t!]
\centering
\small
\begin{tabular}{lr}
\toprule
\textbf{Model} & \textbf{\# Parameters} \\
\midrule
ChatGPT GPT-5 
& ?   \\
ChatGPT GPT-5-nano
& ?   \\
ChatGPT GPT-4 
& ?   \\
Claude Opus-4.6 
& ?   \\
Gemini 2.5-Pro
& ?    \\
Gemini 2.5-Flash-Lite
& ?    \\
Qwen 3-a22b-thinking
& 235B  \\
Mistral Small-Instruct
& 24B  \\
OLMo 3 Instruct
& 7B  \\
\bottomrule
\end{tabular}
\caption{\textbf{Evaluated models.} Parameter counts are listed only when publicly disclosed by the provider. 
}
\label{tab:models}
\end{table*}

\paragraph{Additional Results.} Table~\ref{tab:models}  lists all models included in our evaluation, together with their parameter counts when known. Table~\ref{tab:model_accuracy} reports results for models omitted from our main analysis due to their low performance on the \textit{Muddy Children} puzzle under symmetric inference. All four models score near random chance: although the prompt offers three options (\textit{yes}, \textit{no}, \textit{I don't know}), we evaluate only rounds where the correct answer is determinately \textit{yes} or \textit{no}, placing the effective random baseline at 50\%. Table~\ref{tab:macro_f1} reports macro F1 scores for the main-analysis models across all four benchmark quadrants. In the symmetric setup, the label distribution is 55 Yes and 45 No; in the asymmetric setup, it is 60 Yes and 40 No. For each quadrant, macro F1 is computed separately per puzzle over the two ground-truth classes (\textit{Yes} and \textit{No}), then averaged across puzzles. Since models are queried only at the pivotal round, \textit{I don't know} is never a correct answer and is excluded from the average. While the absolute values naturally differ from the accuracy results in Figure~\ref{fig:success_rates}, as the two metrics capture different aspects of model performance, the macro-F1 scores exhibit the same qualitative patterns: a consistent drop from symmetric to asymmetric settings across all models, and the same relative ordering of behavioral profiles observed in the accuracy results.

\begin{table*}[t!]
\centering
\small
\begin{tabular}{lcccc}
\toprule
\textbf{Model} & \textbf{Classic Sym.} & \textbf{Classic Asym.} & \textbf{Novel Sym.} & \textbf{Novel Asym.} \\
\midrule
Qwen 3-235b-a22b-thinking & 1.0000 & 0.5823 & 0.9974 & 0.6107 \\
Gemini 2.5-Pro & 0.9966 & 0.5032 & 1.0000 & 0.5951 \\
ChatGPT GPT-5-nano & 0.9864 & 0.2787 & 0.9415 & 0.0839 \\
ChatGPT GPT-5 & 1.0000 & 0.3375 & 1.0000 & 0.5616 \\
\bottomrule
\end{tabular}
\caption{\textbf{Macro F1 scores across the four benchmark quadrants.} For each quadrant, macro F1 is computed separately per puzzle over the two ground-truth classes (\textit{Yes} and \textit{No}), then averaged across puzzles. Since models are queried only at the pivotal round, \textit{I don't know} is never a correct answer and is excluded from the average.}
\label{tab:macro_f1}
\end{table*}

\end{document}